\documentclass[10pt,twocolumn,letterpaper]{article}

\usepackage{cvpr}
\usepackage{times}
\usepackage{epsfig}
\usepackage{graphicx}
\usepackage{amsmath}
\usepackage{amssymb}
\usepackage{microtype}

\usepackage[pagebackref=true,breaklinks=true,letterpaper=true,colorlinks,bookmarks=false]{hyperref}

\cvprfinalcopy 

\def\cvprPaperID{****} 

\ifcvprfinal\fi
\begin{document}

\title{\vspace{-\baselineskip}On the use of human reference data for evaluating automatic image descriptions
\\[-\baselineskip]\raisebox{50pt}[0pt][0pt]{\makebox[0pt]{\parbox{\textwidth}{\normalfont\normalsize\centering Originally presented as a (non-archival) poster at the VizWiz 2020 workshop, collocated with CVPR 2020.\\ See: \url{https://vizwiz.org/workshops/2020-workshop/}}}}}

\author{Emiel van Miltenburg\\
Tilburg center for Cognition and Communication (TiCC), Tilburg University\\
Warandelaan 2, 5037 AB Tilburg, The Netherlands\\
{\tt\small C.W.J.vanMiltenburg@tilburguniversity.edu}
}

\maketitle

\section{Introduction}
Automatic image description systems are commonly trained and evaluated using crowdsourced, human-generated image descriptions \cite{bernardi2016automatic}. The best-performing system is then determined using some measure of similarity to the reference data (BLEU \cite{papineni-EtAl:2002:ACL}, Meteor \cite{denkowski:lavie:meteor-wmt:2014}, CIDER \cite{vedantam2015cider}, etc). Thus, both the quality of the systems as well as the quality of the evaluation depends on the quality of the descriptions. 
As Section~\ref{sec:language} will show, the quality of current image description datasets is insufficient. I argue that there is a need for more detailed guidelines that take into account the needs of visually impaired users, but also the feasibility of generating suitable descriptions. With high-quality data, evaluation of image description systems could use reference descriptions, but we should also look for alternatives.

\section{The language of image descriptions}\label{sec:language}
By virtue of their size, current image description datasets such as Flickr30K \cite{young2014image} or MS COCO \cite{lin2014microsoft} provide a unique opportunity to study how people talk about images. Doing so has revealed that:

\begin{enumerate}
\item Despite the fact that the guidelines tell them not to, crowdworkers often speculate about the contents of the images \cite{miltenburg2016stereotyping,vanmiltenburg-elliott-vossen:2017:INLG2017}. The presence of these \textit{unwarranted inference} means that the data does not constitute a reliable basis for evaluation. This work also argues that speculation occurs because of the decontextualized nature of the task. It is hard to describe an image without interpreting it, and interpretating an image often means filling in any missing details.

\item There is a high degree of variation in the descriptions \cite{miltenburg2016stereotyping,miltenburg2018talking,miltenburg2018measuring}. This means that it unclear what descriptions should look like. The diversity in image description datasets is partly due to differences between the annotators \cite{vanmiltenburg-elliott-vossen:2017:INLG2017}, but also due to the undefined nature of the crowdsourcing task, which does not specify what the descriptions will be used for. Desmond Elliott (p.c.) notes that crowdworkers for the German portion of the Multi30K corpus \cite{elliott-EtAl:2016:VL16} actively discussed the purpose of the task on crowdworker forums online. Successful communication requires interlocutors to be aware of the purpose of the exchange, so that they can adjust their contributions accordingly \cite{grice1975logic}.

\item The descriptions contain linguistic constructions, such as negations (e.g.\ \textit{A man \textbf{not} wearing a shirt playing tennis.}) that require high-level reasoning (e.g.\ knowing that people usually wear shirts, and signaling that this behavior is unusual), and as such are impossible to generate for current systems \cite{vanmiltenburg-morante-elliott:2016:VL16}.
\end{enumerate}

These observations raise the question of what image description systems should look like. Although the development of image description guidelines is an active area of research (see \cite{10.1145/3313831.3376404} for an overview), there is a disconnect between the human-computer interaction (HCI) literature and the image captioning (IC) literature. Specifically: (1) the HCI literature does not look at variation in image description data; and (2) on the IC-side, current evaluation practices and image description datasets do not take existing guidelines into account. For the latter, crowdworkers receive fairly generic instructions, which leave the datasets open to (a subset of) the flaws discussed in this section.

\section{Captions for visually impaired people}
To my knowledge, there are currently two image description datasets for blind or visually impaired people. The first is developed by Gella \& Mitchell \cite{gella2016residual}, who interviewed potential users, and developed the Expressive Captions Dataset to address their needs. For this dataset, crowdworkers were explicitly asked to talk about emotional content of the images. Few other details are known, as the dataset remains unreleased.

The second dataset was developed by Gurari and colleagues \cite{gurari2020captioning}, and is currently the target of the VizWiz 2020 image captioning task. A notable improvement from earlier published work, is that the crowdsourcing procedure clearly notes that the description should be useful to someone who is blind. This hopefully lowers the amount of variation in the descriptions, and gives rise to a more uniform dataset. But otherwise, the instructions mostly tell users what \emph{not} to say.
Future work should also provide more positive support.

\section{Conclusion}
Given the above, we should be careful when evaluating image description systems using human reference data. A further complicating factor is that textual similarity-based evaluation metrics have been shown to be unreliable indicators of output quality for natural language generation systems \cite{van-der-lee-etal-2019-best}. So what are we to do? I offer two suggestions:

\begin{enumerate}
    \item Complement user studies with the assessment of human-generated descriptions, to develop more detailed guidelines. This will enable us to more precisely specify the needs of visually impaired users. E.g.\ the question `which features of human entities are/should be described?' has been looked at from both the user \cite{10.1145/3313831.3376404} and data  \cite{miltenburg2018talking} side. Combining these perspectives hopefully means that we do not miss any relevant features.
    \item Develop automatic checks to see whether human- or computer-generated descriptions conform to those guidelines. This might also lead to a ``description-checking interface'' which could provide real-time feedback to users writing image descriptions.
\end{enumerate}

This proposal hopefully brings us closer together, to tailor image description systems to their users' needs.

{\small
\bibliographystyle{ieee_fullname}
\bibliography{bibliography}
}

\end{document}